\documentclass[sn-mathphys-num]{sn-jnl}

\usepackage{hyperref}
\usepackage{graphicx}%
\usepackage{multirow}%
\usepackage{amsmath,amssymb,amsfonts}%
\usepackage{amsthm}%
\usepackage{mathrsfs}%
\usepackage[title]{appendix}%
\usepackage{xcolor}%
\usepackage{textcomp}%
\usepackage{manyfoot}%
\usepackage{booktabs}%
\usepackage{algorithm}%
\usepackage{algorithmicx}%
\usepackage{algpseudocode}%
\usepackage{listings}%

\usepackage{cite}
\usepackage{subcaption}
\RequirePackage{xspace}

\newcommand{\eg}{\emph{e.g}} 
\newcommand{\ie}{\emph{i.e~}}

\newcommand{\objnav}{\textsc{ObjectNav}\xspace}

\newcommand{\moveforward}{\textsc{move\_forward}\xspace}
\newcommand{\movebackward}{\textsc{move\_backward}\xspace}

\newcommand{\turnleft}{\textsc{turn\_left}\xspace}
\newcommand{\turnright}{\textsc{turn\_right}\xspace}
\newcommand{\lookup}{\textsc{look\_up}\xspace}
\newcommand{\lookdown}{\textsc{look\_down}\xspace}
\newcommand{\stopac}{\textsc{stop}\xspace}

\newcommand*{\rowstyle}[1]{
  \gdef\@rowstyle{#1}%
  \@rowstyle\ignorespaces%
}

\theoremstyle{thmstyleone}%
\theoremstyle{thmstyletwo}%

\theoremstyle{thmstylethree}%

\begin{document}

\title[Visual Semantic Navigation with Real Robots]{Visual Semantic Navigation with Real Robots}


\author*[1]{\fnm{Carlos} \sur{Guti\'errez-\'Alvarez}}\email{carlos.gutierrezalva@uah.es}
\equalcont{These authors contributed equally to this work.}

\author[2]{\fnm{Pablo} \sur{R\'ios-Navarro}}\email{pablo.rios@urjc.es}
\equalcont{These authors contributed equally to this work.}

\author[1]{\fnm{Rafael} \sur{Flor-Rodr\'iguez-Rabad\'an}}\email{rafael.flor@edu.uah.es}
\author[1]{\fnm{Francisco Javier} \sur{Acevedo-Rodr\'iguez}}\email{javier.acevedo@uah.es}
\author[1]{\fnm{Roberto Javier} \sur{L\'opez-Sastre}}\email{robertoj.lopez@uah.es}

\affil*[1]{\orgdiv{Signal Theory and Communications}, \orgname{University of Alcalá}, \orgaddress{\city{Alcalá de Henares}, \postcode{28805}, \country{Spain}}}

\affil[2]{\orgdiv{Computer Science}, \orgname{Rey Juan Carlos University}, \orgaddress{\city{Móstoles}, \postcode{28933}, \country{Spain}}}


\abstract{Visual Semantic Navigation (VSN) is the ability of a robot to learn visual semantic information for navigating in unseen environments.
These VSN models are typically tested in those virtual environments where they are trained, mainly using reinforcement learning based approaches.
Therefore, we do not yet have an in-depth analysis of how these models would behave in the real world.
In this work, we propose a new solution to integrate VSN models into real robots, so that we have true embodied agents.
We also release a novel ROS-based framework for VSN, ROS4VSN, so that any VSN-model can be easily deployed in any ROS-compatible robot and tested in a real setting.
Our experiments with two different robots, where we have embedded two state-of-the-art VSN agents, confirm that there is a noticeable performance difference of these VSN solutions when tested in real-world and simulation environments.
We hope that this research will endeavor to provide a foundation for addressing this consequential issue, with the ultimate aim of advancing the performance and efficiency of embodied agents within authentic real-world scenarios.
Code to reproduce all our experiments can be found at \url{https://github.com/gramuah/ros4vsn}.}




\maketitle

\section{Introduction}\label{sec:introduction}

Can a robotic agent navigate and interact in the real world as seamlessly as humans do?
This is the fundamental question driving research within the embodied AI community.
The problem is formally known as Visual Semantic Navigation (VSN), \eg~\citep{ramrakhya2023,Cai2024DGMemLV,chang2020}.
However, mimicking human navigation is a challenging task for robots, particularly in unseen environments, as it requires efficient exploration and a deep understanding of the objects and structures within the space.
For unknown scenarios, humans can leverage prior semantic information achieved from previous scenes to navigate in new environments, but it is still a challenging task to incorporate that knowledge into embodied agents, especially in real robotic platforms navigating in the real world.
The potential of autonomous robots with these advanced navigation capabilities is vast, ranging from assistive robots that can guide individuals with reduced mobility to specific locations, to platforms that can aid in complex environments such as search and rescue operations or logistic centers.

\begin{figure}[t]
    \centering
        \includegraphics[width=\linewidth]{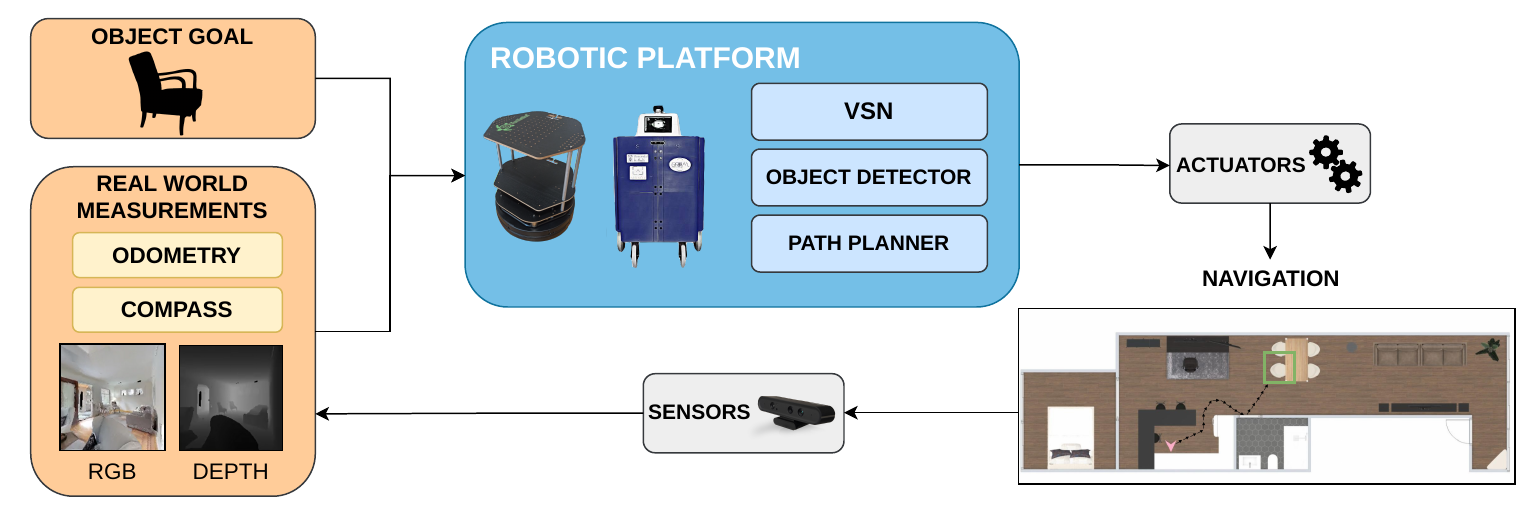}
        \caption{
        \objnav task is a complex navigation problem.
        An agent needs to employ vision-based sensors to navigate from a random starting point to specific object goals within the scene.
        Many different hardware and software components need to be fully integrated to solve it, making it difficult to deploy and test these visual semantic navigation (VSN) models in real robots.
        Therefore, the best current solutions are trained and tested in virtual environments.
        Our goal is to bridge the gap between virtual and physical environments by providing a ROS-based framework that simplifies testing and comparing various VSN models on real robotic platforms.
        }
        \label{fig:abstract}
\end{figure}

Technically, in this work, we focus on embedding VSN models in real robotic platforms.
This is our main objective.
We focus on the Object-Goal Navigation (\objnav)~\citep{batra2020} problem.
As it is seen in figure~\ref{fig:abstract}, in \objnav task the agent has to navigate from a random position to certain object goals present in the scene, mainly using vision-based sensors.
In contrast with traditional geometric navigation approaches, where the navigation problem is solved typically by using a map or generating it on the fly (e.g. SLAM), VSN models are learning-based approaches that use \emph{no} metric map.
Therefore, VSN solutions must learn visual representations of the environment to reduce the exploration time and better generalize to unseen scenes and object categories.
Notably, many of these solutions combine reinforcement learning (RL) and/or imitation learning (IL) strategies with recent advances in deep learning models for visual perception to address the navigation problem in \emph{virtual} environments, where embodied AI agents are trained and tested.
However, we believe it is crucial to thoroughly investigate how the latest solutions to the VSN problem perform when deployed with real robots navigating in real-world environments.
This is where our work makes its most significant contribution.
Although there are VSN works that achieve \textit{near-human} navigation performance, \eg~\citep{ramrakhya2023}, these systems are mainly trained and tested in virtual environments, so how these models would behave in the real world is still a question to answer.

In this work, our objective is to build true embodied agents proposing a new solution for the integration of VSN models into real robots.
The main contributions of our work are as follows:
\begin{enumerate}
 \item We first provide to the embodied agents community the ROS4VSN development, a novel Robot Operating System (ROS)~\citep{ros} based architecture that allows testing and comparing different VSN models in real robots.
 Our ROS4VSN development is model agnostic; thus, any VSN solution can be integrated into it with ease.
 Section~\ref{subsec:ros4vsn} includes all the technical details.
 \item We also have embedded two state-of-the-art VSN models in two different robotic platforms.
 The selected models have been PIRLNav~\citep{ramrakhya2023} and VLV~\citep{chang2020}.
 To achieve this integration, it is necessary to make technical adaptations to the models so that they transition from interacting with observations provided by simulation environments to observations from the real world.
 We detail all these technical modifications to the models in Section~\ref{subsec:vsn_models}.
 \item Finally, we propose an experimental evaluation of the adapted VSN models, using our ROS4VSN, with two different robots, in a real world scenario (Section~\ref{sec:experiments}).
 The main question we want to address with the designed experiments is: Are the state-of-the-art VSN models able to successfully operate with real robots?
 We have measured their success rate in real world navigation experiments, which has allowed us to analyze the difference in performance compared to those tests in simulation environments, where embodied AI agents are typically tested.
 Our work shows that these VSN solutions perform noticeably differently when evaluated in real-world and simulated situations.
 With the ultimate goal of improving the performance and efficiency of VSN systems in real robots, we hope that our work will help to provide the groundwork for tackling this significant challenge.
\end{enumerate}

\section{Related Work}\label{sec:related-work}
\textbf{Visual Semantic Navigation}.
To navigate in unfamiliar environments, traditional methods use depth sensors~\citep{newcombe2011, thrun2001} and RGB cameras~\citep{jones2011, sattler2018} to build geometric maps and simultaneously determine the robot's position in relation to the map.
This is known as Simultaneous Localization and Mapping (SLAM)~\citep{Kazerouni2022, campos2021, labbe2022}).
Typically, these SLAM models use heuristic algorithms to create graph-based representations of the environment, allowing the robot to visit the different nodes of the graph when navigating to specific points.
Semantic SLAM (\eg~\citep{zhang2018, rosinol2020, jin2023}) expands upon SLAM by integrating semantic data from the environment, allowing the robot to identify and store objects in memory.

A recent approach, made possible by advances in machine learning and computer vision, involves designing navigation policies that directly train deep neural networks to learn semantic information from visual observations in an end-to-end fashion (\eg~\citep{ramrakhya2022,yadav2022, gutierrez2019, khandelwal2022, chaplot2020,chang2020}).
This approach is termed \textit{visual semantic navigation} (VSN).
These models often rely on the use of CNNs as visual encoders followed by RNNs; that are in charge of predicting an action distribution directly from raw input observations.
The neural networks are trained using imitation learning (IL) or reinforcement learning (RL) approaches.

When IL is applied to the visual navigation problem, navigation policies are learnt from expert demonstrations (\eg~\citep{ramrakhya2022,yadav2022}).
It can also be used combined with an RL fine-tuning phase to achieve better performance~\citep{ramrakhya2023}.

Other works focus on the use of an end-to-end RL approach to solve \objnav navigation~\citep{zhu2017, gutierrez2019, wijmans2020, khandelwal2022, Liu2022, Yadav2023OVRLV2AS} \hypersetup{citecolor=red}~\citep{Xu2024DeepRL, YokoyamaHM3DOVONAD} \hypersetup{citecolor=blue}.
Some authors have proposed combining the RL training with different strategies, like auxiliary tasks~\citep{ye2021}, improved visual representations via object relation graphs~\citep{yang2018}, semantic segmentations~\citep{Mousavian2018} or combining audio feedback with the visual inputs~\citep{Wang2023, Kondoh2023MultigoalAN}.

Modular-learning based approaches~\citep{chaplot2020, chang2020, skillfusion, Li2023RDDRLAR, zhou2022improving} \hypersetup{citecolor=red}~\citep{Cai2024DGMemLV, Kang2024HSPNavHS, Wang2023ProbableOL, Wasserman2023ExploitationGuidedEF, Yokoyama2023VLFMVF} \hypersetup{citecolor=blue} decompose the navigation process in separate modules that execute different tasks.
It is common for these methods to be composed of a high-level semantic exploration module trained by RL that indicates the agent subgoals that have to be reached by a low-level navigation policy.
Modular learning can be also combined with offline RL~\citep{shah2022} techniques to leverage navigation behaviors from fixed datasets, without any additional online data collection or fine-tuning.

Finally, there are different approaches that try to tackle the problem of rapidly adapting to unseen environments in visual navigation via meta-learning~\citep{wortsman2019, luo2021, zhang2022}.
These methods are trained on a variety of different environments (usually designated as tasks) and are able to generalize to unseen environments by learning a policy that can be quickly adapted to new environments.
And the recent progress in large language models (LLMs) has led to the possibility of using them to solve the visual navigation problem~\citep{Huang2023, Zhou2023} as well.
In this case, the LLMs are used as a reasoning module in charge of understanding the semantic information present on the environment.
They then share this information with different modules in charge of navigating to the specified goal.

Our goal in this work is not to develope VSN approaches, but to integrate various state-of-the-art VSN models into multiple real-world robots by using our novel ROS4VSN library.
Technically, we have chosen to integrate the PIRLNav~\citep{ramrakhya2023} and VLV~\citep{chang2020} models into two different robots.
These integrations required several technical adaptations, particularly in the areas of sensor data integration and navigation planning.
Overall, we are able to show how ROS4VSN allows easily testing and comparing different VSN methods in the real world.
To the best of our knowledge, our work is the first to develop a model agnostic ROS package for visual semantic navigation, where multiple models can be integrated.

\textbf{Simulation-to-reality transfer in robotic navigation}.
Deploying a model trained in simulation to a real robot is a challenging task.
Due to logistical constraints, training a model in the real world —especially with RL techniques— is often impractical, prompting the use of alternative methods to address this challenge
For example, \citet{kim2022} propose a monocular vision-based time-to-collision estimation for small drones by domain adaptation of simulated images.
Their method converts simulated images into real-like synthetic images using a sim-to-real method.
This is done with the aim of minimizing efforts and time invested in the collection of training datasets within real-world scenarios, while simultaneously maximizing the advantages inherent in simulated environments.

Overall, it is necessary to develop methods that allow to efficiently transfer the knowledge learnt in simulation to the real world~\citep{kadian2020}.
Different approaches have been proposed to solve this problem.
For instance, CAD2RL~\citep{sadeghiCAD2RLRealSingleImage2017} system achieved remarkable success in training a collision avoidance policy entirely within a simulated environment.
This breakthrough was subsequently tested on real aerial drones, with promising results.
By focusing on simulation refinement~\citep{Son2020}, the accuracy of simulations can be improved by exploiting the disparities between simulated and real-world observations.
In the field of locomotion, training legged robotic systems in a simulated environment and subsequently transferring the acquired policies to real-world applications~\citep{Hwangbo_2019, agarwal2022} has always been a challenging task.

For the problem of VSN, we have the study by \citet{gervet2022} that shows how their approaches perform in real-world settings.
However, we would like to highlight the novel contributions that our work offers.
First, while~\citep{gervet2022} focuses mainly on the comparison of their navigation methods, we here, along with a similar study, release to the research community the modular ROS4VSN software architecture.
Our main goal is to facilitate the prototyping of new VSN solutions on real robots.
So, we offer a ROS-compatible software architecture, model agnostic, that allows a simple integration of different VSN approaches in ROS robots.
In this way, future VSN solutions will be able to be tested on real robots in a convenient and straightforward manner.
Second, we include in our study more recent VSN solutions than the ones reported in~\citep{gervet2022}, as the PIRLNav model~\citep{ramrakhya2023}, which defines the state-of-the-art for the \objnav problem.
Third, we also provide, for the first time, a detailed analysis on how a model directly trained with real videos, such as the VLV~\citep{chang2020}, performs in real robots.
This allows us to compare, as in~\citep{gervet2022}, how a modular-learning model (\ie VLV~\citep{chang2020}), compares with a typical end-to-end learning approach (\ie PIRLNav).
Interestingly, our study also concludes, like in~\citep{gervet2022}, that modular-learning approaches perform better in the real world.
Fourth, in our study, we employed two different robotic platforms: one commercially available and widely used by various laboratories, and another custom-built.
This demonstrates the versatility of the proposed solution, showing that it can be integrated into different robots.
And finally, in our work, we propose an experimental evaluation specifically designed for testing in the real world, which can be employed in future research studies.
Overall, we hope that our ROS-based library will help to further advance the field of visual semantic navigation in real robots.

\section{Methodology}\label{sec:methodology}
In this research work, our main objective has been to efficiently integrate various state-of-the-art VSN models on multiple robotic platforms.
To fulfill our goals, the first step of the proposed methodology has been to develop a ROS-based solution to ease the integration of VSN solutions into real robots.
It is crucial that the approach is model-agnostic, allowing the integration of any VSN-designed model.
We have named this development ROS4VSN, and it is detailed in Section~\ref{subsec:ros4vsn}.
Once the ROS4VSN system is available, we need to select some state-of-the-art VSN models that will allow us to experimentally evaluate them with real robots.
In Section~\ref{subsec:vsn_models}, we justify the selected VSN models and detail the technical modifications made to integrate them into ROS4VSN\@.


\subsection{ROS4VSN: ROS for Visual Semantic Navigation}
\label{subsec:ros4vsn}

\begin{figure}
    \centering
    \includegraphics[width=\linewidth]{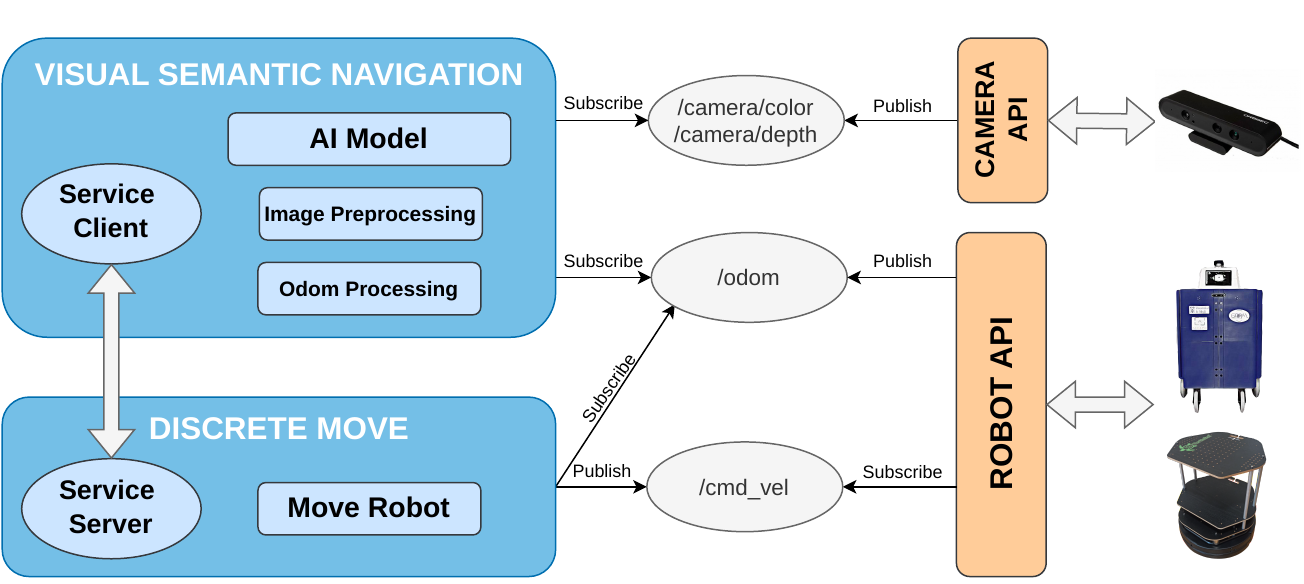}
    \caption{Architecture scheme of the ROS44VSN framework.
    It shows the different packages, topics and connections within them and the hardware devices.}
    \label{fig:arch_scheme}
\end{figure}

We have designed ROS4VSN library to be modular and flexible, so that it can be easily adapted to different robots and VSN models.
It is built on top of ROS \textit{Noetic}~\citep{ros} open-source robotic middleware, because of its flexibility, support, compatibility and popularity among the robotics community.
ROS provides a collection of useful tools, libraries, and conventions to simplify the task of creating complex and robust robot behaviors across a wide variety of robotic platforms, which makes it perfect for our framework.
We design the architecture of the framework, so it has three main capabilities: it can receive and process information from the environment, infer actions using an AI VSN model, and control the actuators of a platform to reach a specific navigation goal.
It makes it easy to integrate different VSN models, since it only needs to replace the model with which the experiments are to be carried out.
The architecture is divided into the following main packages, each of which plays a specific role: robot\_api, camera\_api, discrete\_move and visual\_semantic\_navigation.
These packages are connected to each other through ROS topics and services, and also to external hardware devices: a camera (RGB + Depth) and a differential drive robotic platform.
Figure~\ref{fig:arch_scheme} shows a visual representation of the global architecture scheme, illustrating the connections between the different developed ROS packages and the hardware devices.

\subsubsection{Robot API}\label{subsubsec:robot-api2}

This package is responsible for controlling the actuators of the robot and sending odometry information to the discrete\_move and visual\_semantic\_navi\-gation packages (depicted in figure~\ref{fig:arch_scheme}).
It is typically designed by the manufacturer of the robot, so it can be different depending on the employed platform.
In our particular case, the development of the framework and experiments were done using two robots (see Figure~\ref{fig:robots}).
First, a Turtlebot 2 robot, so the standard turtlebot2~\citep{kobuki} package is integrated as robot\_api.
Since the Turtlebot 2 expects the velocity commands in the \textit{/mobile\_base/commands/velocity} topic, we perform a remapping to the \textit{/cmd\_vel} topic from our discrete\_move package.
Our second robot is known as LOLA2~\citep{LOLA}.
We have developed its complete robot\_api package to guarantee the compatibility with the rest of ROS4VSN architecture.

In charge of the communication with the platform, this package is also responsible for publishing the robot's odometry information through the topic \textit{/odom} (odometry topic).
This information is crucial for the visual\_seman\-tic\_navigation and discrete\_move packages.
On one hand, the package discrete\_move uses this information to adjust the velocity commands sent to the robot\_api package, achieving precise and controlled movements.
On the other hand, the visual\_semantic\_navigation package can use the odometry information to help infer the action to be executed by the robot or to help a planner reach its destination.

\subsubsection{Camera API}\label{subsubsec:camera-api}

This package is responsible for capturing RGB and depth images from the camera.
It publishes them through the \textit{/camera/color} and \textit{/camera/depth} topics, respectively.
The camera used in our robots is the Orbecc Astra S, an RGB plus depth camera, based on structured light technology.
We have adapted the official ros\_astra\_camera package~\citep{orbeccros} to be integrated in our ROS4VSN architecture.
The modular design of ROS4VSN allows it to be used with any other type of camera on the robots, simply by adapting this Camera API package.

\subsubsection{Discrete Move package}\label{subsubsec:discrete-move-package}

\begin{figure}
    \centering
    \includegraphics[width=\linewidth]{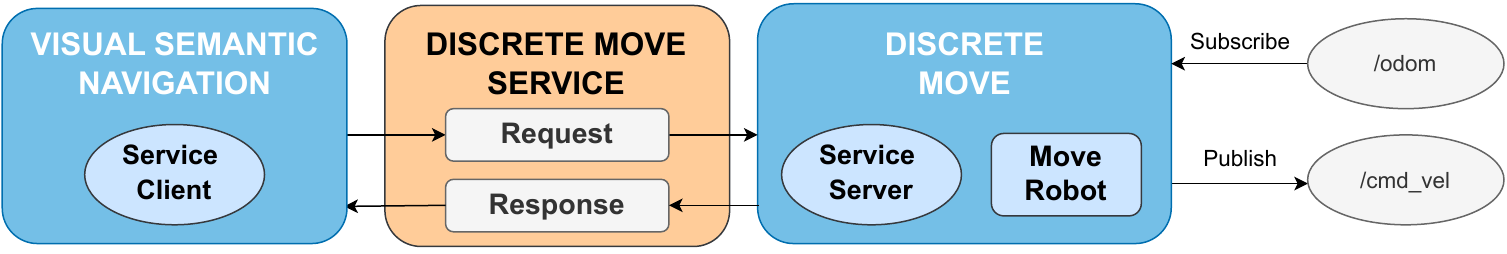}
    \caption{Communications between visual\_semantic\_navigation and discrete\_move packages.}
    \label{fig:discrete_move}
\end{figure}

This package has been developed with the purpose of providing a precise and customizable control of the robotic platform through discrete navigation commands.
Note that this is the way most VSN models interact with their agents (\eg~\citep{ramrakhya2023,chang2020}).
In embodied AI, navigation in simulated environments is performed using discrete action commands that agents execute to reach the specific goals: move forward (25 cm), turn left or right (30 degrees), or stop.
Our ROS4VSN package acts as a server to which clients can request a set of discrete movements and configure the forward distance or turning angles.
The package communications scheme is shown in detail figure~\ref{fig:discrete_move}.
It is in charge of executing the actions requested by the visual\_semantic\_navigation package and sending a response when the action is completed.

\paragraph{\textbf{Set of navigation movements}}\label{par:movement-set}

The set of movements allowed by the package consists of the following actions: \turnleft, \turnright, \moveforward, \movebackward and \stopac.
All the actions are fully customizable in terms of distance and angle, except for the \stopac action, which does not require any additional parameters since it just stops the robot.
This package has been designed as a ROS service, so the communication between the visual\_semantic\_navigation package and the discrete\_move package is done synchronously and bidirectionally.
That way, the visual\_semantic\_navigation package can wait for the response of the discrete\_move package when the action has been completed before sending any new action request.
The discrete\_move server is in charge of sending the right \textit{/cmd\_vel} commands to the robot\_api package, so the robot can execute the requested action and receive the \textit{/odometry} topic information from the robot\_api package.
That way, it can calculate the movement done by the robot, and stop the action when the requested action has been finished.

Embodied AI navigation environments, such as Habitat~\citep{NEURIPS2021_021bbc7e}, are simulation environments where there are no movement errors in the agents.
However, our scenario is the real world, with real robots.
Therefore, ROS4VSN must integrate error control strategies.
To achieve this, the discrete\_move package includes two error correction strategies: one for the turn error $\epsilon_{turn}$ and one for the move straight error $\epsilon_{straight}$.

The turn error is calculated in degrees as follows,
\begin{equation}
    \label{eq:turn_error}
    \epsilon_{turn} = (\alpha_{target}- \alpha_{current}) \bmod(360)\; ,
\end{equation}
where $\alpha_{target}$ is the target orientation and $\alpha_{current}$ is the current orientation of the robot.

The move straight error is computed as:
\begin{equation}
    \label{eq:error_move}
    \epsilon_{straight} = {d - \sqrt{(x-x_{init})^2+(y-y_{init})^2}}\; ,
\end{equation}
where $d$ is the displacement distance requested by the system, $(x,y)$ encodes the current position of the robot and $(x_{init},y_{init})$ defines the initial position of the robot.
We consider a successful turn when $\epsilon_{turn}$ is less than 0.1 degrees and a successful displacement when $\epsilon_{straight}$ is less than 5 millimeters.
Our ROS4VSN architecture continuously measures these errors to adapt the rotation and displacement movements of the robots, ensuring that they occur with the highest possible precision.

\paragraph{\textbf{Acceleration and braking control}}\label{par:start-and-brake-control}

When developing a navigation system based on discrete commands, it is crucial to implement appropriate braking and acceleration mechanisms to achieve smooth and efficient robot navigation.

The package includes an implementation that combines a constant acceleration until the desired maximum speed is reached, with a deceleration phase to stop the robot.

\begin{figure}
    \centering
    \includegraphics[width=\linewidth]{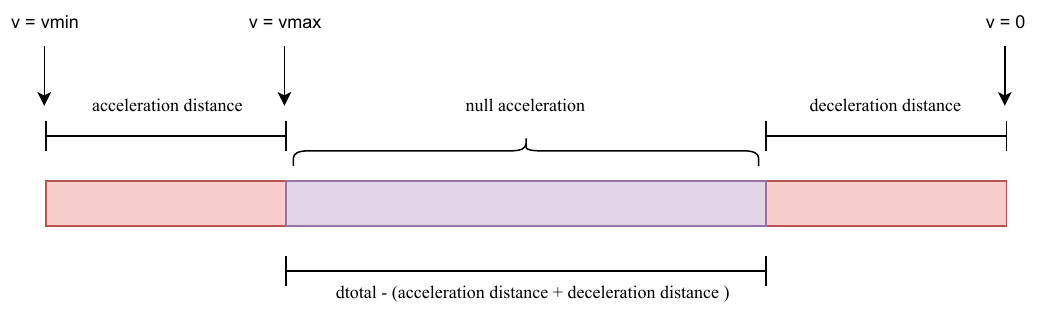}
    \caption{Acceleration and braking control scheme.}
    \label{fig:acceleration_stop}
\end{figure}

The smoothness of the movement and the time needed to complete it depend on the percentage of the path in which the robot is accelerating and decelerating, as well as on the initial speed.
By properly adjusting these parameters, a smoother movement and a more efficient navigation time can be achieved.
Figure~\ref{fig:acceleration_stop} illustrates the total distance that the robot must travel for a \moveforward (or \movebackward) command.
This distance is divided so that the robot performs the following phases: acceleration, displacement at constant speed and deceleration.
The instantaneous speed of the robot is controlled so that during the first phase, there is a uniformly accelerated motion, according to the following equation: $v = \sqrt{v_{init}^2 + 2 a \epsilon_{straight}}$, where $a$ is the desired acceleration, $v_{init}$ represents the initial speed of the robot, and $\epsilon_{straight}$ encodes the distance covered by the robot.
Note that we must continuously read $\epsilon_{straight}$ using our ROBOT API, in order to dynamically adjust the speed.
For the deceleration stage, we employ an equivalent negative acceleration in the previous equation.
With these equations, we can progressively adjust the linear speed ($v$) that is sent to the platform to obtain a smooth navigation.

\paragraph{\textbf{Configuration parameters}}\label{par:configuration-parameters}

The discrete move package includes the configuration file \texttt{discrete\_move.yaml}, where the different parameters used for the execution can be modified.
Table~\ref{tab:discrete_configuration} shows the configuration parameters and its default value.

\begin{table}
    \centering
    \begin{tabular}{c|c}
        \toprule
        \textit{\textbf{Parameter}}             & \textit{\textbf{Default Value}} \\
        \midrule
        Linear Velocity                         & 0.3 m/s                         \\

        Angular Velocity                        & 0.5 rad/s                       \\

        Acceleration and deceleration distances & (\moveforward distance) / 3     \\
        \bottomrule
    \end{tabular}
    \caption{Configuration parameters of discrete\_move package.}
    \label{tab:discrete_configuration}
\end{table}

\subsubsection{Visual Semantic Navigation package}\label{subsubsec:visual-semantic-navigation-package}

One of the main features of our ROS4VSN software is that it allows to easily integrate different VSN models independently of the robotic platform used.
To achieve this, it is essential to be agnostic with respect to the software environment needed by the particular VSN model, as each model may require different dependencies.

The goal of this package is to simplify the deployment of VSN models on real robots, providing an efficient software structure for the execution of these methods.
In other words, it aims to facilitate the inference tasks of discrete movement actions that these systems produce, using the state of the robotic platform.
The state is defined by the robot's position in the real world, the information provided by its sensors, and the action that was previously taken.

VSN models make decisions using mainly RGB images of the environment.
However, some of them can also use additional information, such as the position and orientation of the robot, or even depth images.
Therefore, this package must be responsible for: a) capturing all the information from the sensors and processing the obtained data; b) inferring with the VSN models the next navigation action; and c) communicating with the robotic platform to request the corresponding discrete motion.
This process is repeated iteratively, collecting new data, making inferences and executing actions, until a stop condition is reached or an error occurs.

In our architecture, the package connects to the camera through the \textit{/camera} topic published by the camera\_api module, to receive the necessary RGB and depth images.
It also collects information from the robot's odometry through the \textit{/odometry} topic published by the robot\_api package.
Furthermore, it acts as a client of the discrete\_move package to send the actions determined by the VSN model and receive confirmations about their execution.
This package contains two additional submodules: image preprocessing and odom processing.


\paragraph{Image Preprocessing submodule}\label{par:image-preprocessing}

This submodule is in charge of collecting and preprocessing the images from the camera of the robot.
These images are necessary for the VSN model to infer the appropriate action.
The package must be able to communicate with the camera in real time and receive its information.
This communication is done through the \textit{/camera/color} and \textit{/camera/depth} ROS standard topics.
Typically, depth images are taken using a time-of-flight camera.
This type of camera can lead to noise problems, including incomplete data in certain areas of the image, noise on metallic surfaces, and the impact of scene lighting on distance measurements.
To address these problems, a temporal median filter is implemented, so for a series of $N$ depth images, its noise can be reduced by discarding outliers.

\paragraph{Odom Processing submodule}\label{par:odom-processing}

This submodule is in charge of collecting odometry information by subscribing to the \textit{/odom} topic (\textit{odometry topic}), published by the robot\_api package.
This odometry information consists of two main variables: 1) robot position, that indicates the current location of the robot with respect to its initial position; and 2) robot orientation, that defines the current direction in which the robot is oriented in relation to its initial orientation.
Some state-of-the-art VSN models (\eg~\citep{ramrakhya2023}) need to input these two sources of information.

\paragraph{\textbf{Module workflow}}\label{par:module-workflow}

Our VSN package uses the image submodule and the odometry submodule to capture the camera images and the robot odometry data.
This information can then be passed as input to a particular VSN model.
Using the VSN model integrated in the package, the next navigation action that the robot must execute is inferred.
Once the action has been determined, the package sends a message (\textit{request}) to the discrete\_move server to request the execution of the movement by the robot.
After sending the request to the discrete\_move server, the package waits to receive a confirmation message.
This message indicates whether the requested action has been performed correctly or if some problem has occurred.
If for some reason an action has not been completed successfully, the server has been programmed to return $False$, which completely stops the execution of the workflow.
It is important to highlight that the workflow is repeated until whether the \stopac action is inferred by the model, the time limit for the episode is reached or the server responds with a message indicating some problem during the execution.

A configuration file is provided (\texttt{vsn.yaml}) containing default values for the parameters of our VSN package.
By modifying this file, one can easily change the navigation target, the parameters associated with the median filter, or even the maximum number of steps allowed to be executed during a navigation exercise.


\subsection{VSN models}\label{subsec:vsn_models}
For this research work, we have decided to adapt and integrate into our robots two state-of-the-art VSN models: PIRLNav~\citep{ramrakhya2023} and VLV~\citep{chang2020}.
The first model, known as PIRLNav~\citep{ramrakhya2023}, is a VSN approach that has been trained with a combination of imitation learning and a RL fine-tuning.
As of today, this model reports the best results in the \objnav~\citep{batra2020} task in Habitat~\citep{NEURIPS2021_021bbc7e}.
The second model is the VLV approach~\citep{chang2020}, which is a VSN model directly trained from YouTube videos.
The VLV model makes use of such videos to learn semantic cues for an effective navigation to semantic targets in indoor scenarios.
VLV is a modular learning solution that combines low-level and high-level navigation policies.

These models are complementary in the sense that they are based on two paradigms: a) imitation learning plus RL; and b) modular learning.
This aspect will allow us to study, in the experimental evaluation, which type of approach yields better results in the real world.
Note that we do not intend to retrain these models but rather subject them to evaluation in the real world.
We aim to analyze their generalization ability for navigation outside simulation environments.
It is precisely thanks to our ROS4VSN system that this can be done, as the technical modifications made to the models will be oriented towards embedding them in a ROS-based system.
Next, we provide a detailed description of the modifications and adaptations made to these models so that they can be integrated into ROS4VSN and tested on real robotic platforms.

\subsubsection{VLV}

The first approach is known as Value Learning from Videos (\textsc{VLV}), developed by \citet{chang2020}.
VLV is a modular-learning based VSN model directly trained from videos of real state agencies, taken from YouTube.
In this type of video, a human records, camera in hand, the properties for sale, showcasing all the rooms they have, to generate a sort of virtual tour of the houses.
Note that the videos used do not have any type of information about the navigation actions that take place during the recording, nor in what kind of rooms or what type of objects appear.

The VLV model leverages such YouTube videos to learn semantic cues for an effective navigation to semantic targets in indoor home environments.
This way, the VLV model is trained to find in these videos a set of object categories.
Technically, the model uses pseudo action labels obtained by running an inverse model on the navigation sequences.
This inverse model is able to recognize the discrete movements that each of the transitions of the different video frames involve.
Then, the navigation policies are learned following a reinforcement learning (RL) approach.
VLV employs Q-learning to learn from the video sequences that have been pseudo-labeled with the actions.
The learned Q-function, and the associated value function, implicitly learn semantic cues for navigation.
In other words, the model learns what images lead to the desired category and what do not.

\begin{figure}
    \centering
    \begin{subfigure}[b]{\textwidth}
        \includegraphics[width=\textwidth]{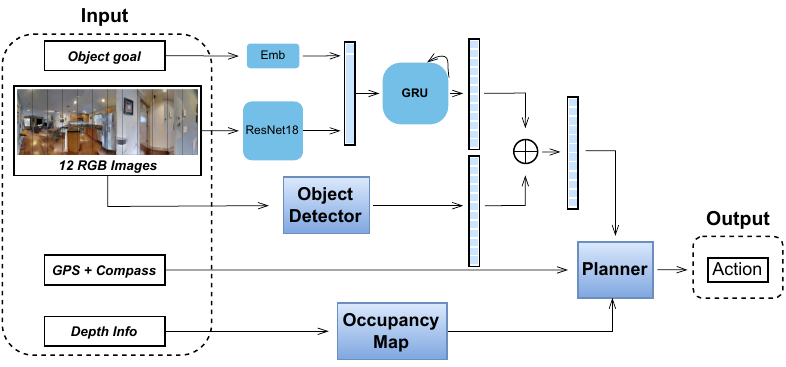}
        \caption{VLV model.}
        \label{fig:vlv_overview}
    \end{subfigure}
    ~
    \begin{subfigure}[b]{\textwidth}
        \includegraphics[width=\textwidth]{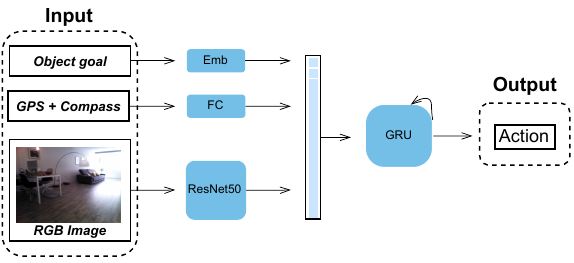}
        \caption{PIRLNav model.}
        \label{fig:pirlnav_overview}
    \end{subfigure}
    \caption{VSN models integrated into our VSN-ROS.}\label{fig:vsn_models_overview}
\end{figure}

For our experiments, we had to embed the VLV model in our ROS4VSN architecture.
See Figure~\ref{fig:vlv_overview}.
Technically, we integrated the two navigation policies detailed in the experiments in~\citep{chang2020} that were tested in the virtual environment Habitat~\citep{NEURIPS2021_021bbc7e}.
However, now, our goal is to implement them in a real robot in the real world.

This integration into our robots, using our ROS4VSN architecture, has consisted of the following steps.
First, a high-level policy that stores 12 images for each node in a topological graph (obtained by rotating 12 times by 30 degrees each) is used.
This high-level policy uses the learned value function score over these 12 images, and samples the most promising direction for seeking objects of a particular object category.
VLV needs an object detector output to produce the final score for these images.
This way, the high-level policy is equipped with a mechanism to seek the object once it has been detected.
Specifically, the detector we employ is the Mask R-CNN~\citep{mask-rcnn}, which we had to embed in our architecture as well.
To navigate, this policy will use the following discrete movements/actions: \moveforward 25cm, \turnright $30^\circ$ or \turnleft $30^\circ$.

These movements are compatible with the developed discrete\_move package of our ROS4VSN architecture.
Once a main direction has been chosen, our approach converts it into a short-term goal by sampling a location at an offset of 1.5 meters from the chosen node\textquotesingle s location, in the chosen view\textquotesingle s direction.
This is done using the depth-camera and a low-level navigation policy that uses occupancy maps with a fast marching planning~\citep{Sethian1996} to execute robot actions to reach the short-term goal.
These two policies have been integrated into our ROS4VSN nodes, and with them, we get the robot to explore the environment.

\subsubsection{PIRLNav}
The second model selected for our experimental evaluation is known as \textsc{PIRLNav}~\citep{ramrakhya2023}.
As of today, this model reports the best performance in the \objnav~\citep{batra2020} task in Habitat~\citep{NEURIPS2021_021bbc7e}.

\begin{figure}[t]
    \centering
    \includegraphics[width=0.6\linewidth]{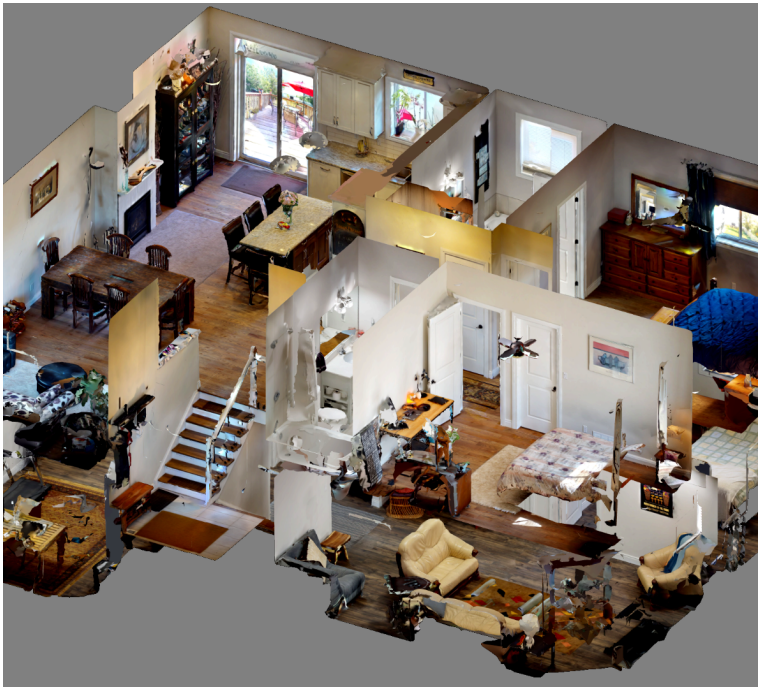}
    \caption{One of the 3D reconstructions of the HM3D-Semantics v0.1 dataset~\citep{Ramakrishnan2021HabitatMatterport3D}, used to train PILRNav.}
    \label{fig:scene_hm3d}
\end{figure}

PILRNav is a VSN approach that has been trained with a combination of imitation learning and a RL fine-tuning.
The model uses behavior cloning (BC) to pre-train the \objnav policy on a dataset of $77k$ human demonstrations, amounting more than $2370$ human annotation hours, in the HM3D-Semantics v0.1 dataset~\citep{Ramakrishnan2021HabitatMatterport3D}.
This dataset provides up to 120 different 3D reconstructions of houses all around the world (see Figure~\ref{fig:scene_hm3d} for an example of one of them).
Once this BC is finished, a RL fine-tuning is used following the DD-PPO approach~\citep{Wijmans2019DDPPOLN}.
The policy architecture used is a simple CNN plus an RNN from~\citep{yadav2022}.

In order to integrate PIRLNav into our ROS4VSN architecture, we had to perform the following actions.
See figure~\ref{fig:pirlnav_overview}.
The original PIRLNav needs to receive as inputs the RGB image that the agent observes, as well as the noiseless GPS and compass information offered by Habitat simulator.
GPS and compass Habitat sensor provide the agent\textquotesingle s current location and orientation information relative to the start of the episode.
In our case, because PIRLNav has to be integrated in a real robot, navigating in the real world, we proceed to feed the model with the RGB images that are acquired by the cameras in our robotic platforms.
GPS information is obtained through the odometry information provided by the robot.
For the compass, we recover the relative orientation analyzing all the robot's turning movements.
Note that these are not anymore noiseless sensors, as the ones used in the simulated world in which the PIRLNav model was trained.
Fortunately, we did not observe any important impact on the performance of the model, due to this loss of precision for these sensors.

PIRLNav is therefore integrated in the ROS4VSN architecture to control de navigation of the robot as it has been detailed.
For every captured image, as well as the GPS+compass data, the model is able to determine the next discrete movement action to be executed by our robotic platforms.
The action space used for our experiments with this model is: \moveforward 25 cm, \movebackward 25 cm, \turnright $30^\circ$, \turnleft  $30^\circ$ and \stopac.
The original PIRLNav was also trained to produce the discrete action \lookup and \lookdown, since the simulated agent could tilt its camera.
However, as it is observed in Figure~\ref{fig:robots}, in our platforms these actions are not possible.
We decided to replace \lookup with a \movebackward action, and \lookdown movement with the \moveforward action.
This choice is based on the reasoning that, by raising the camera, more of the scene is captured; moving the robot backward serves this purpose.
Also, since lowering the camera provides a greater level of scene detail, moving forward is considered the most appropriate choice to replace the \lookdown action.
Finally, to prevent collisions between the robot and objects in the scene, a procedure was developed that uses information from the depth image to detect obstacles at a given distance.
Note that PIRLNav does not need any low-level policy as in the VLV model.
The robot is controlled and navigates using only the set of discrete actions provided by the ROS4VSN model.

\section{Experiments}\label{sec:experiments}
This section describes the experimental evaluation designed for testing our developments in the real world.
The goal of our experimental evaluation is to answer the following question: Are the state-of-the-art VSN models able to successfully operate with real robots?
We start with a detailed description of the experimental setup, where the experimental conditions and the evaluation metric are explained.
We then follow with an analysis of the results obtained in the real world.

\subsection{Experimental setup}
\label{subsec:experimental_setup}

One of the primary objectives of our work is to provide a comprehensive and clear protocol for the experimental evaluation of state-of-the-art VSN models in real-world scenarios using real robots.
Our goal is for other researchers to carry out similar experimental evaluations, using the same evaluation metrics, to facilitate comparisons of how different VSN models perform in real-world navigation tasks.
For doing so, we propose the following experimental setup.

\begin{figure}[t]
    \centering
        \includegraphics[width=\linewidth]{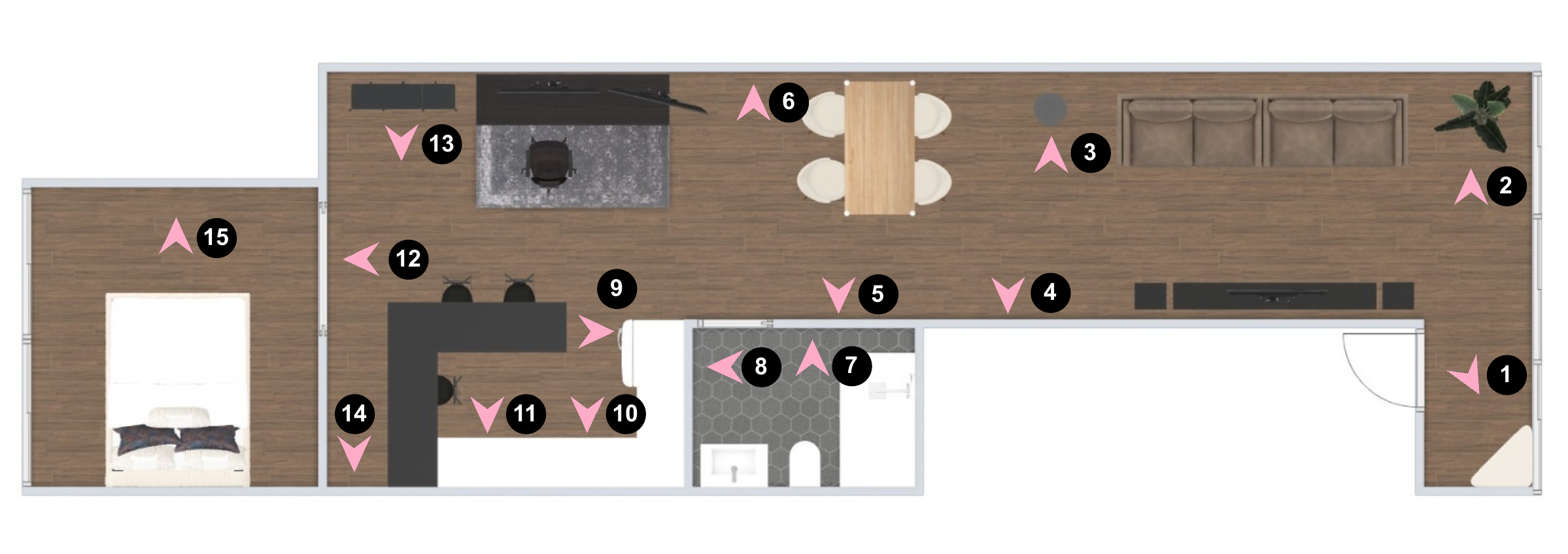}
        \caption{Floor plan where the experiments were performed, indicating the 15 starting positions used.}
        \label{fig:floor_plan}
\end{figure}

In a 75 m$^2$ apartment, we define up to 15 random starting positions (see Fig.~\ref{fig:floor_plan}).
The apartment can be divided into three main areas: a bedroom, a bathroom, and a larger space that includes the kitchen, living room, and study area.
This setting contains all the object categories used to train the VSN models in the experiments, such as chair, bed, plant, bathroom, monitor, table, and sofa.
We encourage other researchers to conduct their experiments in real-world settings that are similar in size and characteristics, allowing the robot to navigate from at least 15 different starting positions across multiple instances.

From these positions, the robot is tasked with navigating to various object categories.
Consequently, one must conduct 15 navigation experiments for each target category and measure the success of the episodes based on whether the robot reaches the designated object category in fewer than 150 discrete actions and without any collisions.
This limit of 150 steps was chosen to establish a balance between the average size of the houses typically used in Habitat~\citep{NEURIPS2021_021bbc7e} and the apartment used in our experiments.

For the evaluation metric, we propose reporting the success rate (SR) of the VSN models as the percentage of episodes in which navigation is deemed successful.
An experiment is considered successful if the robot halts (when the VSN model samples the action \stopac) and the Euclidean distance to the target object is less than one meter.

Note that our navigation experiment mimics the evaluation performed in the \objnav~\citep{batra2020} task, with the same metric.
This is the standard experiment on which most VSN models are compared and which currently defines the state of the art.

\begin{figure}
    \centering
    \begin{subfigure}[b]{0.4\textwidth}
        \includegraphics[width=\textwidth,trim=3cm 0cm 3cm 0cm, clip]{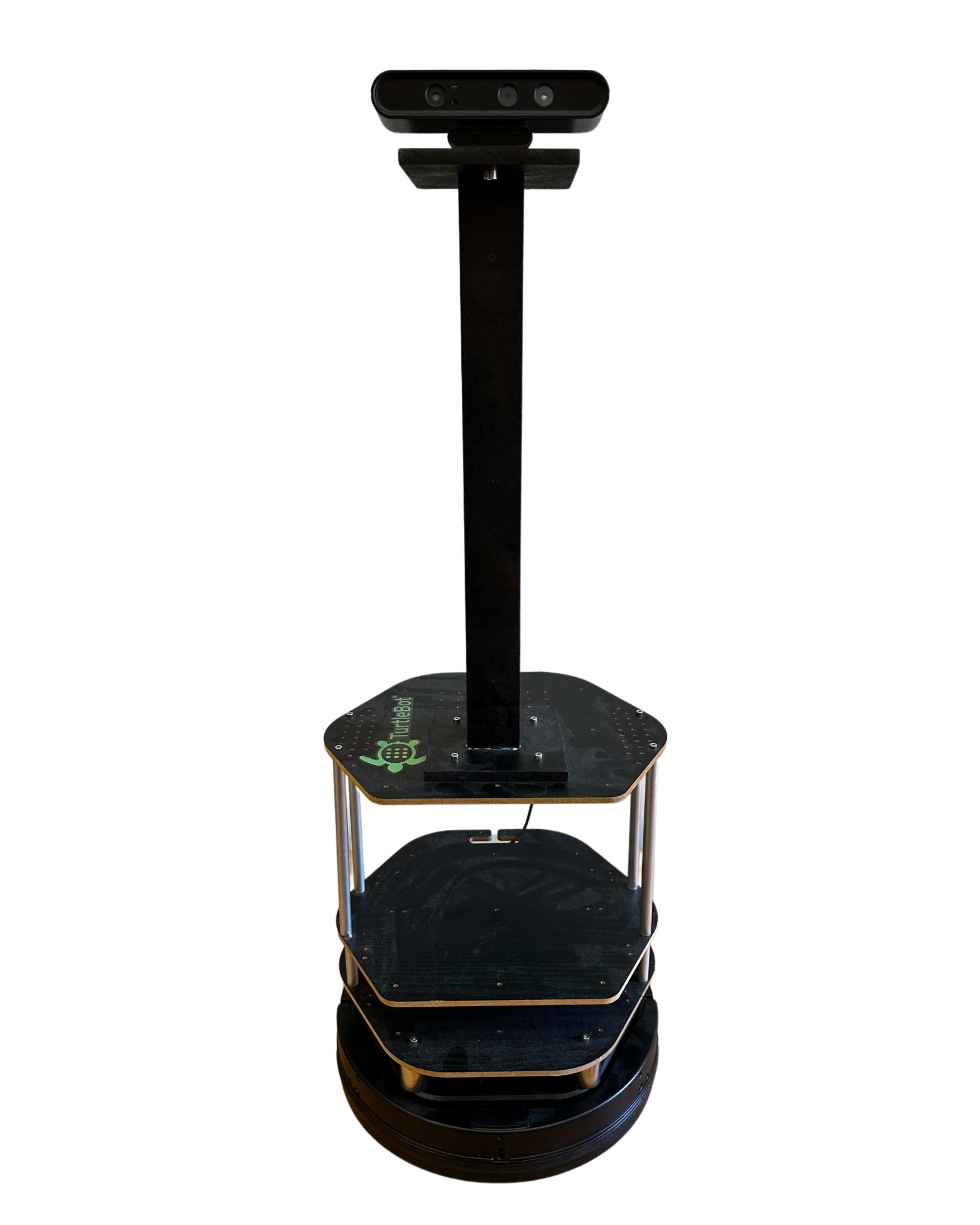}
        \caption{Turtlebot 2.}
        \label{fig:robot_turtlebot}
    \end{subfigure}
    ~
    \begin{subfigure}[b]{0.4\textwidth}
        \includegraphics[width=\textwidth]{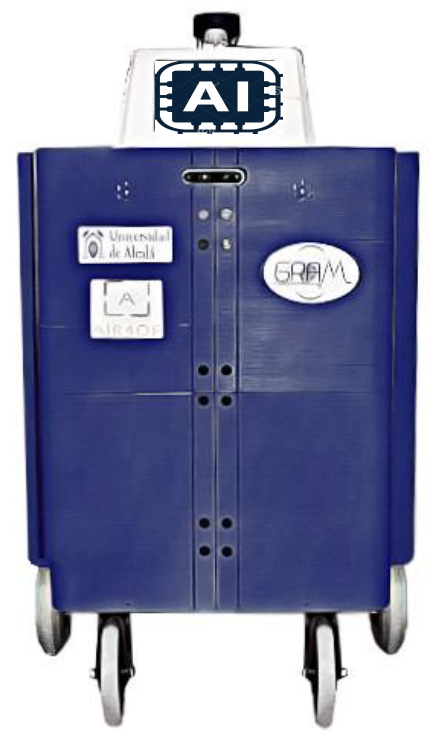}
        \caption{LOLA2 platform.}
        \label{fig:robot_lola}
    \end{subfigure}
    \caption{Pictures of the robots used in our experiments.}\label{fig:robots}
\end{figure}

We have used two different robots for our experiments: a Turtlebot 2, and the LOLA2~\citep{LOLA} platform.
To do so, we had to embed our ROS4VSN in both of these platforms.
This can be easily done by adapting the robot\_api module described in Section~\ref{subsubsec:robot-api2}.
Figure~\ref{fig:robots} shows the family picture of the robots invited to our experiments.
We mainly used the Turtlebot 2 for the navigation experiments in the apartment described.
The robot LOLA2 was also used in navigation experiments, but in a different location, to test the stability of the developed system and to provide a study that contains more hours of navigation, and on different platforms.
Our intention in using two different platforms has been to provide evidence of the generalizability of our architecture, demonstrating that it can be tested on different robots.

When collecting information during the experiments, we developed a procedure to record relevant data for each trial.
This procedure stores the unique identifier for each episode, the sequence of actions performed, and the category of object searched.
In addition, during the tests, qualitative information about the trajectory followed by the robot was recorded.
In particular, all the images observed by the robot during its trajectories have been saved.

\begin{figure}
    \centering
        \includegraphics[width=\linewidth]{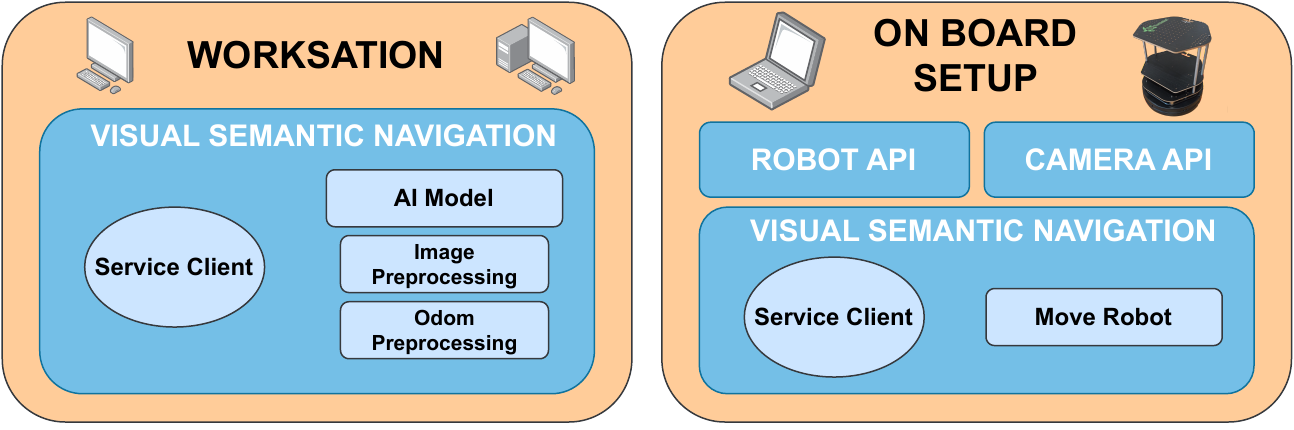}
        \caption{Hardware-software architecture for the development of experiments.}
        \label{fig:setup_experiment}
\end{figure}

For the experiments, the following hardware-software setup has been used, as it is shown in Figure~\ref{fig:setup_experiment}.
The modular architecture of the developed ROS4VSN system was used to deploy it in a distributed manner.
This architecture allows separating the execution of ROS packages on different devices, as long as they are connected to the same network.
The robotic platforms were equipped with a laptop.
This device was used to establish the communication with the robotic platform and the camera by executing the robot\_api and camera\_api packages.
At the same time, the discrete\_move package was executed to receive the actions to be executed by the robot.
On the other hand, we used a workstation to run the VSN nodes, which was equipped with an i7-1165G7 processor and an NVIDIA GeForce RTX 2060 graphics card.
We provide code to reproduce all our experiments at \url{https://github.com/gramuah/ros4vsn}.

\subsection{VSN navigation results}
\label{subsec:vsn}
We detail in this section the main results obtained during the navigation experiments with our robots, including both quantitative and qualitative results.

Following the experimental setup detailed in Section~\ref{subsec:experimental_setup}, we have obtained the following results.
Remember that we provide in this study an analysis of the SR for two state-of-the-art VSN models, running in a Turtlebot 2 platform.
For every VSN model, \ie VLV, and PIRLNav, we have measured their SR for the different object categories they can navigate to.

Table~\ref{tab:results_sota} compares the performance obtained by VLV and PIRLNav approaches when they are tested in the real world (\ie our experiments) and in a virtual environment (\ie the experiments reported in their respective papers).
The first thing we observe is the difference in terms of SR\@.
The SR for the PIRLNav model drops from $65\%$ to $21\%$, while the VLV model loses $\sim10$ percentage points in this metric.
One of the conclusions of our study is that there is a considerable gap between the behavior of these models in the real world and in the simulation environments in which they are trained.
This indicates that further research in this direction is needed.
Interestingly, the results obtained in our real-world experiments are not consistent with the performance difference that already existed between the models in the simulation environments: VLV is the winner in the real world!
As we analyze in the discussion section below (See section~\ref{subsec:discussion}), we believe that this behavior is due to the impact of the object detector that VLV integrates, but PIRLNav does not.
While the difference between VLV and PIRLNav SR in the virtual environments is of $26$ percentage points, in the real world this gap becomes of just $8$ percentage points.

\begin{table}
    \centering
    \begin{tabular}{l|cc}
        \toprule
        \textbf{Models} & SR (Real World) & SR (Virtual Environment) \\
        \midrule
        \textsc{VLV}~\citep{chang2020} & $29.33\%$ & $39\%$ \\
        \textsc{PIRLNav}~\citep{ramrakhya2023} & $21.11\%$ & $65\%$ \\
        \bottomrule
    \end{tabular}
    \caption{Real world success rate against simulation.}
    \label{tab:results_sota}
\end{table}

In the following, we analyze in detail the results reported by each of the models.
We start with the VLV model.
Table~\ref{tab:vlv} reports the SR for every of the target categories used in our experiments.
Chairs, tables, and sofas are the categories that are easiest to navigate to.
In analyzing various trials with the robot using the VLV model with ROS4VSN, there were no successful outcomes from starting positions 10 and 12 (see Figure~\ref{fig:floor_plan}).
Additionally, only one success was observed from positions 1, 2, and 13.
Notably, our VLV implementation can reach most targets in under 60 steps.

\begin{table}[t]
\centering
\begin{tabular}{c|cccc}
\toprule
\textit{\textbf{Object Goal}} & \textit{\textbf{Successful episodes}} & \textit{\textbf{SR}} &  \textit{\textbf{Avg. number of actions}}   \\ \midrule
Chair                & 6/15     & 40\%   &   30  \\
Sofa                 & 6/15     & 40\%   &   65  \\
Table                & 6/15     & 40\%   &   42  \\
Bed                  & 3/15     & 20\%   &   39  \\
Toilet               & 1/15     & 6,67\%    &   42  \\ \bottomrule
\end{tabular}
\caption{VLV VSN experiment. We report the number or successful episodes over 15, the corresponding SR per-object goal, and the average number of actions taken to reach the target.}
\label{tab:vlv}
\end{table}

\begin{figure*}[t]
    \centering
        \makebox[\textwidth][c]{
        \includegraphics[width=1.2\linewidth]{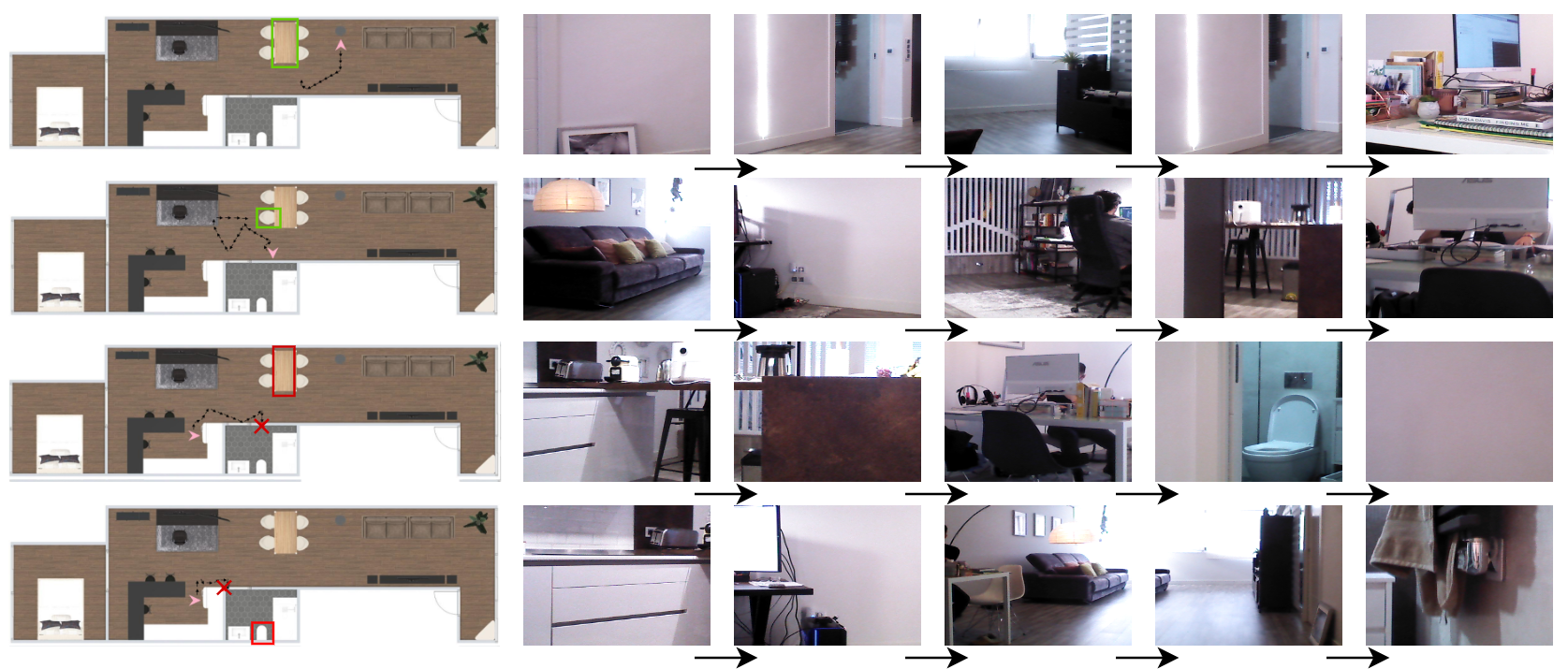}}
        \caption{VLV qualitative navigation results. The first two rows show two successful cases, where the robot reached the target, while the last two rows show two situations where the navigation experiment failed.}
        \label{fig:vlv_qualitative}
\end{figure*}

Figure~\ref{fig:vlv_qualitative} shows qualitative results for four navigation experiments with the VLV model.
We provide five representative images of the navigation experiments.
Two successful and two unsuccessful cases are presented.
In the first experiment (first row), the robot quickly reached the Table, as the detector easily identified it in the images.
In the second experiment, the robot took a detour to the Chair because the detector failed to detect it initially.
The model predicted a point near the chair but out of view, causing the robot to move closer only after it became visible.
In the third and fourth experiments, the robot started in a challenging position in front of a refrigerator.
Due to noise in the depth image, the target point calculation was inaccurate, leading to collisions with the wall in both cases.

We analyze now in detail the results reported by PIRLNav model.
Table~\ref{tab:pirlnav} shows the SR obtained for the PIRLNav model integrated into our ROS4VSN architecture.
We detail the SR reported for every object category.
The agent was able to more easily locate the most common objects in the house, such as the chair and the monitor.
The abundant and well-distributed presence of these objects facilitated the agent's work.
The model inferred the action \stopac on both objects a total of five times.
Large objects, such as the sofa and bed, showed slightly lower results.
Although easily visible from multiple locations in the dwelling, the presence of only one of these objects made it difficult for the robotic agent to spot them.
The number of times the \stopac action was sampled for these categories was substantially reduced.
With the toilet, we have one of the most complex challenges in navigating the robot in this home.
The robot did not have full visibility of the toilet until it managed to fully enter the bathroom, having to pass through the narrow door without colliding.
Finally, for the category plant, the robot was not able to locate this category in any of the 15 attempts.
Even though the plant was visible on multiple occasions during navigation, the agent did not manage to head towards this object.
Overall, considering all the categories, the SR for the PIRLNav model is of $21.11\%$.

\begin{table}[t]
\centering
\begin{tabular}{c|cccc}
\toprule
\textit{\textbf{Object Goal}} & \textit{\textbf{Successful episodes}} & \textit{\textbf{SR}} &  \textit{\textbf{Avg. number of actions}}   \\ \midrule
Chair                & 5/15     & 33,33\%   &   49  \\
Monitor              & 5/15     & 33,33\%   &   91  \\
Sofa                 & 5/15     & 33,33\%   &   70  \\
Bed                  & 3/15     & 20,00\%   &   97  \\
Toilet               & 1/15     & 6,67\%    &   61  \\
Plant                & 0/15     & 0,00\%    &   82  \\ \bottomrule
\end{tabular}
\caption{PIRLNav VSN experiment. We report the number or successful episodes over 15, the corresponding SR per-object goal, and the average number of actions taken to reach the target.}
\label{tab:pirlnav}
\end{table}

\begin{figure}
    \centering
        \makebox[\textwidth][c]{
        \includegraphics[width=1.2\linewidth]{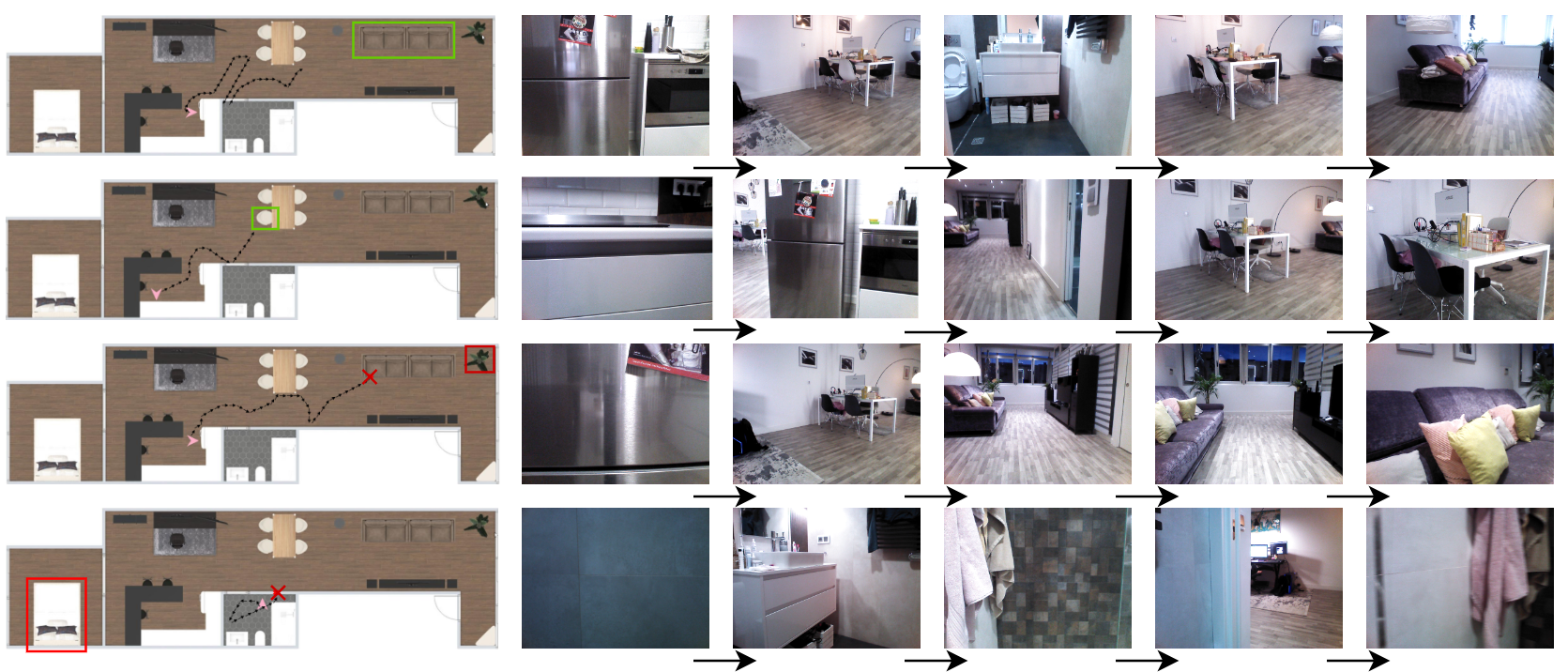}}
        \caption{PIRLNav qualitative navigation results. The first two rows show two successful cases, where the robot reached the target, while the last two rows show two situations where the navigation experiment failed.}
        \label{fig:pirlnav_qualitative}
\end{figure}

To conclude our analysis of the PIRLNav model, we provide some qualitative results.
Figure~\ref{fig:pirlnav_qualitative} shows the navigation trajectories for four different experiments.
We provide five representative images of the navigation experiments.
Two successful and two unsuccessful cases are presented.

In the first experiment, the robot starts from the refrigerator and navigates through the house until it reaches the sofa.
This episode is carried out in 69 actions and ends when the model infers the action \stopac in front of the sofa.
The second experiment is also a success story, but this time the robot starts navigating from the kitchen.
The robot leaves the kitchen and navigates to the nearest chair.
This episode is performed in 36 actions and ends with the \stopac action determined by the model.
The third experiment shows a case of navigation failure, where the robot targeting the plant hits an obstacle.
In this episode, the robot navigates for 61 actions until it hits the couch.
Despite visualizing the plant from far away, when trying to approach it, the robot ends up crashing.
In the last episode, another case of navigation failure is shown while the robot was trying to make its way to the bed.
As it can be seen in this episode, the model had difficulty getting through the bathroom door without hitting itself.

We provide a video (see \url{https://youtu.be/nD0JBWNCMGg}) with more qualitative results for both of the VSN models used in our experiments.

\subsection{Stability analysis}
\label{subsec:marathon}

In this section, we analyze the stability of the navigation solution proposed, showing how it can robustly navigate a considerable distance, in two different robots and over two different scenarios.
The robots successfully navigated over more than 5 kilometers in less than 38 hours in two dynamic environments.
The robots operated without direct assistance throughout the experiments, being automatically operated by the VSN models integrated into our ROS4VSN software architecture.

\begin{table}
    \centering
    \begin{tabular}{l|cc}
        \toprule
        \textbf{Robot} & Time (Hours) & Distance (km) \\
        \midrule
        LOLA2       & 8     & 1.12 \\
        Turtlebot 2 & 30    & 4.10 \\\midrule
        Total       & 38    & 5.22 \\
        \bottomrule
    \end{tabular}
    \caption{Time spent and traveled distance for both robots during the experiments.}
    \label{tab:stability}
\end{table}

\subsection{Discussion}
\label{subsec:discussion}
The main question we wanted to address with the designed experiment has been: are the state-of-the-art VSN models able to successfully operate with real robots?
This implies knowing the SR that these models are capable of delivering when tested in the real world, and not in the virtual environments where they were trained.
Note that we selected two VSN models that were originally trained with images of the real world.
Our intention was to reduce as much as possible the influence of domain shift, which we know affects artificial intelligence systems.
Our study confirms that there is still room for improvement so that these models can achieve the same SR in real robots.
We expect, therefore, that our ROS4VSN library plays a fundamental role in this line of research.

Analyzing the particular behavior of the models, we can provide the following interesting discussion.
The integration of the VLV and PIRLNav models within our ROS4VSN architecture has proven to be successful.
It has resulted in a mobile agent capable of navigating in closed environments autonomously, obtaining many experiments where the robotic platforms reach the target class without complications and following logical and direct trajectories.
This navigation is comparable to that observed within simulated environments.
For the VLV model that integrates an object detector, we have observed that this fact has a significant impact on the agent's navigation, especially when it is close to the target class.
Although the object detector does not significantly affect the general exploration, its impact becomes crucial when the robot is in the vicinity of the target.
At this crucial stage of navigation, the object detector provides a significant advantage by guiding the robot to the target more effectively.
This explains the difference of performance we have observed between VLV and PIRLNav in the real world.

\begin{figure}[t]
    \centering
        \includegraphics[width=0.6\linewidth]{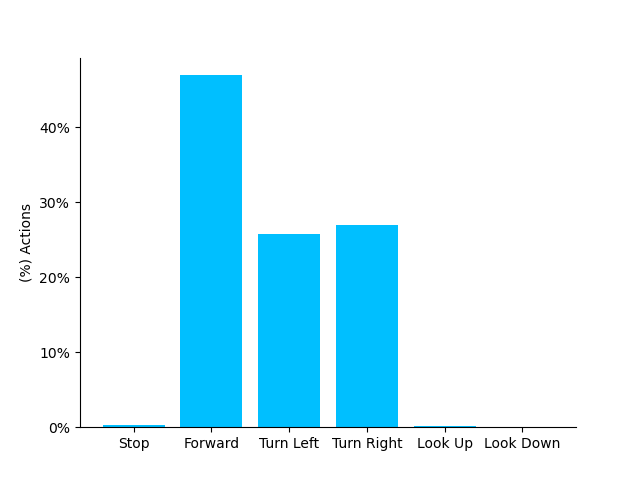}
        \caption{Histogram of navigation actions sampled by PIRLNav model.}
        \label{fig:histrogram_pirlnav}
\end{figure}

In terms of qualitative aspects of navigation, we believe that the PIRLNav model is better than the VLV model.
Note that VLV, every time the high-level policy has to make a decision, needs the robot to turn completely on itself, taking 12 captures on which it will decide which direction to move forward.
This can be observed in the provided video.
This feature slows down navigation, although it could be solved with some specific hardware.
In contrast, PIRLNav offers a more direct navigation experience.
It is interesting to observe the type of action sampling that PIRLNav performs while being executed on our robots.
Figure~\ref{fig:histrogram_pirlnav} shows a histogram corresponding to the distribution of navigation actions performed by PIRLNav.
First, one can observe that the \lookup and \lookdown actions have hardly been selected.
This allows us to affirm that the impact of the adaptations we have made to replace these actions by backward and forward movements, respectively, could hardly have had a considerable impact on the final results.
Second, the most popular actions, as they are the ones that motivate the exploration of the environment, are those of advances and turns.
The stop action was sampled 31 times.
This is a $0.2\%$ as it is reflected in the histogram provided.
We believe that work should be done on solutions to increase the number of times the stop action is selected, but to do so reliably.

Finally, our study confirms some of the conclusions reported in recent works, \eg~\citep{gervet2022}.
Modular-learning models, such as VLV, perform better than end-to-end learning approaches, \eg~PIRLNav, when tested in the real world.

\section{Conclusions}\label{sec:conclusions}
To conclude, we have presented a ROS-based framework for visual semantic navigation named ROS4VSN that allows to easily test and compare different VSN models in real robots.
Using ROS4VSN, we have been able to embed two cutting-edge VSN models into two distinct real robotic platforms.
The chosen models are PIRLNav~\citep{ramrakhya2023} and VLV~\citep{chang2020}.
To seamlessly integrate these models, technical modifications have been needed.
These adaptations ensure a smooth transition for the models, enabling them to shift from interacting with observations generated in simulation environments to those obtained from the real world.
We have also offered a thorough experimental evaluation to showcase how these VSN approaches behave when navigating in the real world.
Our novel framework shows a robust stability, being able to run for a considerable distance, in two different robots, without any human intervention.
Our study and results show that the performance of state-of-the-art VSN models is significantly lower in the real world than in the virtual environments where they were trained.
We expect that our efforts will lay the foundation for addressing this significant challenge.

\section*{Acknowledgements}
This research was partially funded by projects: NAVISOCIAL, with reference 2023/00405/001 from the University of Alcal\'a; NAVIGATOR-D, with reference PID2023-148310OB-I00 from the Ministry of Science and Innovation of Spain.\\
\newpage

\bibliography{library}

\end{document}